\documentclass{article}

\usepackage[preprint]{neurips_ai4d3_2025}

\usepackage[utf8]{inputenc} 
\usepackage[T1]{fontenc}    
\usepackage{hyperref}       
\usepackage{url}            
\usepackage{booktabs}       
\usepackage{amsfonts}       
\usepackage{nicefrac}       
\usepackage{microtype}      
\usepackage{xcolor}         
\usepackage{amsmath}
\usepackage{graphicx}       
\usepackage{float}          
\usepackage{subcaption}     

\title{Transcoder-based Circuit Analysis for Interpretable
Single-Cell Foundation Models}

\author{%
  Sosuke Hosokawa \\
  The University of Tokyo\\
  Tokyo, Japan \\
  \texttt{hoso-sosuke0421@g.ecc.u-tokyo.ac.jp} \\
  \And
  Toshiharu Kawakami \\
  The University of Tokyo Hospital \\
  Tokyo, Japan \\
  \texttt{kawakami-toshinaru555@g.ecc.u-tokyo.ac.jp} \\
  \And
  Satoshi Kodera \\
  The University of Tokyo Hospital \\
  Tokyo, Japan \\
  \texttt{koderasatoshi@gmail.com} \\
  \And
  Masamichi Ito \\
  The University of Tokyo Hospital \\
  Tokyo, Japan \\
  \texttt{mitou.tky@gmail.com} \\
  \And
  Norihiko Takeda \\
  The University of Tokyo Hospital \\
  Tokyo, Japan \\
  \texttt{ntakeda-tky@g.ecc.u-tokyo.ac.jp} \\
}

\begin{document}

\maketitle

\begin{abstract}
  Single-cell foundation models (scFMs) have demonstrated
  state-of-the-art performance on various tasks, such as cell-type
  annotation and perturbation response prediction, by learning gene
  regulatory networks from large-scale transcriptome data. However, a
  significant challenge remains: the decision-making processes of
  these models are less interpretable compared to traditional methods
  like differential gene expression analysis. Recently, transcoders
  have emerged as a promising approach for extracting interpretable
  decision circuits from large language models (LLMs). In this work,
  we train a transcoder on the cell2sentence (C2S) model, a
  state-of-the-art scFM. By leveraging the trained transcoder, we
  extract internal decision-making circuits from the C2S model. We
  demonstrate that the discovered circuits correspond to real-world
  biological mechanisms, confirming the potential of transcoders to
  uncover biologically plausible pathways within complex single-cell models.
\end{abstract}

\section{Introduction}

In recent years, single-cell foundation models (scFMs) such as
cell2sentence (C2S) \cite{Levine2024C2S} and Geneformer
\cite{Theodoris2023} have garnered significant
attention in the field of computational biology. These models adapt
techniques from large language models (LLMs) in natural language
processing, combining pre-training on large-scale transcriptome data
corpora to learn general gene-gene relationships with task-specific
fine-tuning on smaller datasets \cite{Theodoris2023,Cui2024}. While
these models have achieved
state-of-the-art performance on various single-cell analysis tasks
including cell-type annotation and cellular response prediction,
their low interpretability, stemming from the inherent nature of
neural network algorithms, remains a significant challenge. This is
particularly crucial in single-cell analysis models, where biological
interpretation of model predictions is essential, making improved
interpretability in scFMs an urgent priority.

A major goal in efforts to elucidate the internal mechanisms of
large-scale models like LLMs and scFMs includes identifying internal circuits:
the combinations of features that determine model behavior (circuit
tracing) \cite{transcoder_circuits_nips2024,Elhage2021Framework}. In
single-cell analysis models, discovered internal
circuits could potentially lead to new discoveries when connected
with biological insights. This pursuit of understanding model
internal mechanisms falls under the research field of mechanistic
interpretability, which has recently attracted considerable attention.

In the domain of natural language processing, mechanistic
interpretability of LLMs has emerged as a major research topic, with
various methods being proposed. Among these, sparse autoencoders
(SAEs) \cite{huben2024sparse} and their variant, transcoders
\cite{transcoder_circuits_nips2024}, have gained attention as
methods that can resolve the "polysemanticity" within LLMs and
transform internal representations into interpretable features.
Transcoders, in particular, have been shown to extract
input-invariant and highly interpretable features by training neural
networks with wide, sparsely activated intermediate layers that
replace MLP layers, enabling the extraction of model internal circuits.

\begin{figure}[h]
  \centering
  \includegraphics[width=0.9\textwidth]{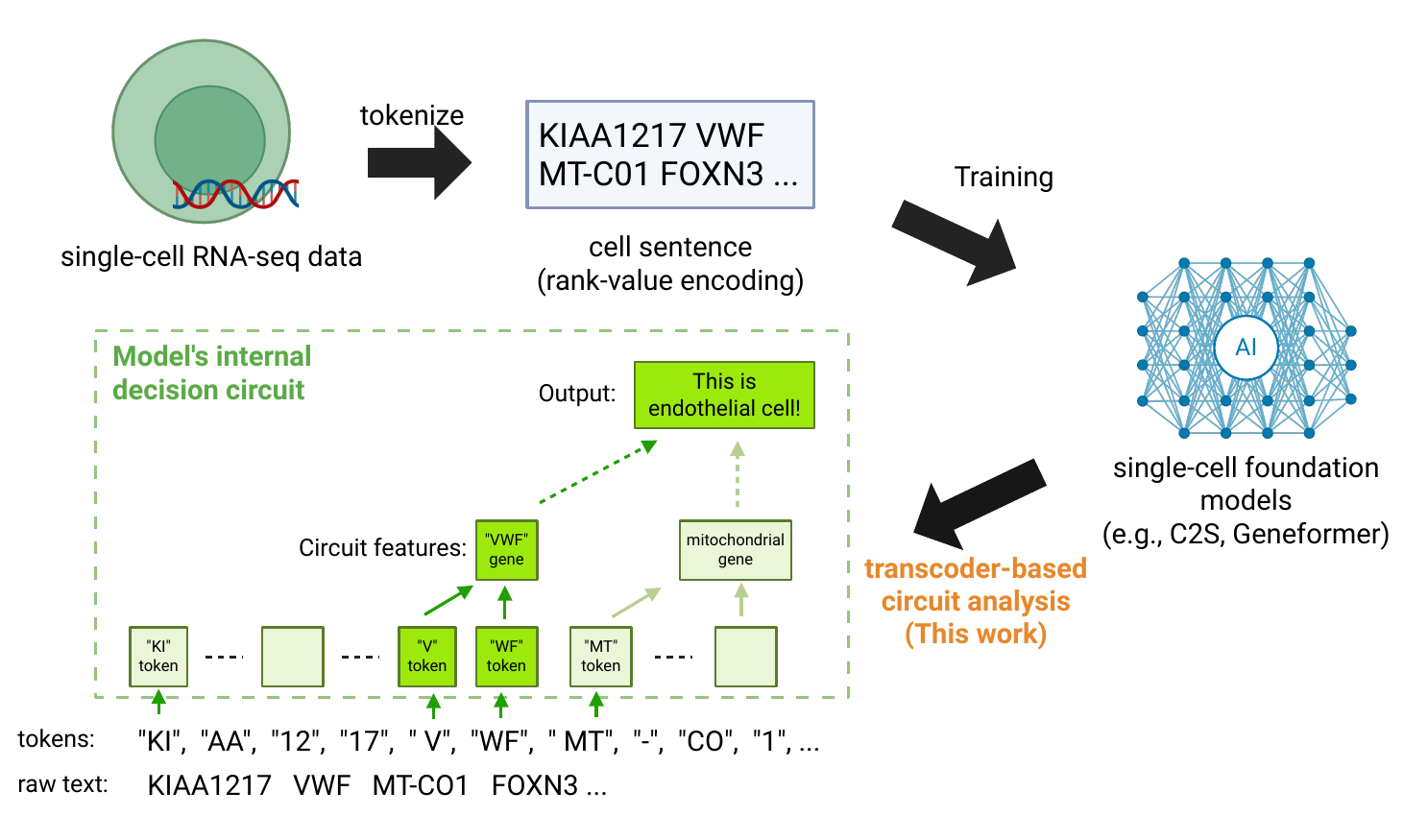}%
  \caption{Pipeline for transcoder-based circuit tracing in scFMs. We train a transcoder on each MLP and attribute across features and attention to recover a sparse computational subgraph (circuit) that underlies cell-type predictions. The recovered features align with known endothelial biology (e.g., VWF, PTPRB, SPARCL1).}%
  \label{fig:overview}%
\end{figure}

In this work, we apply transcoders to the C2S model, a
state-of-the-art scFM, to extract its internal circuits and
biologically interpret the circuit components (Figure
\ref{fig:overview}). Our contributions are
summarized as follows:
\begin{itemize}
  \item We apply transcoders to scFMs (for the first time to our
    knowledge), demonstrating their effectiveness for mechanistic
    interpretability of scFMs.
  \item We establish correspondences between extracted circuit
    components and biological knowledge, showing that biologically
    plausible circuits can be extracted.
  \item We discuss challenges and future directions when applying
    transcoders to scFMs for mechanistic interpretability.
\end{itemize}

The remainder of this paper is organized as follows. We first explain
the background methods of scFMs and transcoders along with their use
for circuit analysis. We then discuss case studies of experiments
applying transcoders to the C2S model. Subsequently, we summarize
related work on interpretability in single-cell analysis models, and
finally present conclusions and future perspectives.

\section{Single-cell Foundation Models}

Single-cell foundation models (scFMs) are transformer-based models
pre-trained on large-scale transcriptome data. These models process
input by ranking genes within each cell based on their expression
levels and other factors, then arranging genes in rank order to form
gene sequences. Prominent examples include Geneformer, scGPT, and
cell2sentence (C2S).

ScFMs exhibit several architectural variations:
\begin{itemize}
  \item \textbf{Architecture type}: Models can be either
    encoder-based or decoder-based.
  \item \textbf{Tokenization method}: Some models tokenize at the
    gene level, while others leverage natural language tokenizers to
    process gene sequences represented as natural language strings.
  \item \textbf{Gene ranking methods}: Different approaches exist for
    ranking genes \cite{Theodoris2023,Levine2024C2S}, which
    represents a unique challenge specific to scFMs.
\end{itemize}

In this work, we focus on the C2S model, which employs a
decoder-based architecture and utilizes natural language
tokenization. C2S leverages the Pythia \cite{Biderman2023Pythia}
architecture and tokenizer,
pre-trained on 57 million human and mouse cells from scRNA-seq data,
along with biological literature abstracts \cite{Levine2024C2S}. This
approach enables the
model to capture both gene expression patterns and broader biological
knowledge from scientific texts.

\section{Transcoders and Circuit Tracing}

\subsection{The Need for Transcoders: Resolving Polysemanticity}

Understanding the internal representations of LLMs faces a
fundamental challenge known as \textit{polysemanticity}: the
phenomenon where individual neurons or weights simultaneously encode
multiple distinct concepts or functions \cite{Elhage2022ToyModels}.
For instance, a single
neuron might strongly respond to both Japanese city names and gene
names. This polysemanticity makes it difficult to disentangle and
understand which parts of the weight matrices represent specific
concepts through direct observation.

To resolve or mitigate polysemanticity and reconstruct the internals
of large models into human-interpretable units, methods such as
sparse autoencoders (SAEs) and their variant, transcoders, have
proven effective \cite{huben2024sparse,transcoder_circuits_nips2024}.

\subsection{Sparse Autoencoders and Transcoders}

\subsubsection{Sparse Autoencoders (SAE)}

SAEs consist of an encoder that transforms an input vector
$\mathbf{x} \in \mathbb{R}^d$ into a higher-dimensional activation
vector $\mathbf{z} \in \mathbb{R}^l_{\geq 0}$ (where $l > d$), and a
decoder that reconstructs the original dimension:
\begin{align}
  \mathbf{z} &= \mathrm{ReLU}(\mathbf{W}_{\mathrm{enc}} \mathbf{x} +
  \mathbf{b}_{\mathrm{enc}}) \\
  \hat{\mathbf{x}} &= \mathbf{W}_{\mathrm{dec}} \mathbf{z} +
  \mathbf{b}_{\mathrm{dec}}
\end{align}

Typical SAEs use the same LLM hidden state for both encoder input and
decoder output, and are trained to minimize the following loss function:
\begin{equation}
  \mathcal{L} = || \hat{\mathbf{x}} - \mathbf{x} ||_2^2 + \lambda
  ||\mathbf{z}||_1
\end{equation}
where the first term represents reconstruction error, the second term
is a sparsity penalty that encourages sparse activations, and
$\lambda$ is a hyperparameter controlling the L1 weight.

\subsubsection{Transcoders}

Transcoders are a variant of SAEs that learn on the input and output
of each transformer layer's MLP rather than on the same hidden state:
\begin{equation}
  \mathcal{L} = || \hat{\mathbf{x}} - \mathrm{MLP}(\mathbf{x}) ||_2^2
  + \lambda ||\mathbf{z}||_1
\end{equation}

This formulation enables transcoders to approximate the transformer's
MLP, decomposing MLP neurons into interpretable components.

\subsubsection{Key Differences between SAEs and Transcoders}

While standard SAEs are trained to reproduce hidden states and
extract input-dependent features, transcoders approximate specific
modules within transformers (the MLPs) and thus extract
input-invariant features. For explaining general model behavior,
input-invariant features are preferable; therefore, transcoders are
better suited for extracting circuits within transformers.

\subsection{Circuit Tracing with Transcoders}

Recent work has proposed methods for tracing circuits within LLMs
using transcoders. We outline the key components below.

\subsubsection{Attribution Between Transcoder Feature Pairs}

The contribution of transcoder feature $i$ in layer $l$ to feature
$j$ in layer $l'$ is computed as:
\begin{equation}
  z^{(l,i)}(x) \times \big(f^{(l,i)}_{\mathrm{dec}} \cdot
  f^{(l',j)}_{\mathrm{enc}}\big)
\end{equation}
where $z^{(l,i)}(x)$ represents the input-dependent activation level,
and the dot product $(f_{\mathrm{dec}} \cdot f_{\mathrm{enc}})$ is an
input-independent fixed value. Here,
$f^{(l,i)}_{\mathrm{enc}/\mathrm{dec}}$ denotes feature vector $i$ of
the transcoder encoder/decoder in layer $l$, corresponding to row
vectors of $\mathbf{W}_{\mathrm{enc}}$ and column vectors of
$\mathbf{W}_{\mathrm{dec}}$ respectively. This decomposition allows
separate treatment of "input-independent general connections" and
"input-specific importance."

\subsubsection{Attribution Through Attention Heads}

Inter-feature relationships propagate not only through MLPs within
the same token but also from different tokens via attention heads.
Through the OV matrices of attention heads \cite{kamath2025tracing},
we can track which
token's information contributes to specific features. Mathematically,
by combining attention scores with OV matrices, we can compute how
representations from source tokens contribute to downstream transcoder features.

\subsubsection{Finding Computational Subgraphs}

By iteratively applying the attribution calculations above, we can
identify the primary computational paths that activate specific
features \cite{transcoder_circuits_nips2024}. The process involves:
\begin{enumerate}
  \item Search for upstream features that strongly contribute to the
    target feature
  \item Retain only top contributors and extend the paths
  \item Iterate to obtain a set of important computational paths
\end{enumerate}
Integrating these paths yields a sparse \textit{computational
subgraph} (circuit) that represents the model's internal
decision-making process.

\section{Experiments}

\subsection{Training Transcoder on C2S}

\subsubsection{Experimental Setup}

We trained transcoders on each MLP layer of
\texttt{vandijklab/C2S-Pythia-410m-cell-type-prediction}
\cite{HF_C2S_Pythia_410m_CellTypePred}, a model from the C2S family
from Hugging Face. For training data, we used the Heart Cell Atlas v2
\cite{heart_cell_atlas_v2} dataset, splitting it into 90\% for
training and 10\% for validation.

\subsubsection{Training Hyperparameters}

The transcoder training was conducted with the following hyperparameters:
\begin{itemize}
  \item Maximum learning rate: $1 \times 10^{-4}$
  \item Number of tokens per batch: 2048
  \item L1 coefficient: $1.4 \times 10^{-4}$
  \item Hidden dimension of transcoder: 8192 (expansion factor 8)
  \item Number of training tokens: 60 million
\end{itemize}

\subsubsection{Model Validation}

To validate the trained transcoders, we compared three models: the
original model, a model with all MLPs replaced by transcoders, and a
model with all MLPs removed. Table \ref{tab:validation_loss} shows
the validation losses for each model configuration.

\begin{table}[h]
  \centering
  \begin{tabular}{lccc}
    \toprule
    Model Configuration & Original & Transcoder & No MLP \\
    \midrule
    Validation Loss & 2.48 & 4.63 & 12.67 \\
    \bottomrule
  \end{tabular}
  \caption{Validation loss comparison across different model configurations.}
  \label{tab:validation_loss}
\end{table}

While the transcoder-replaced model shows some degradation compared
to the original model, it achieves substantially lower loss than the
model with MLPs removed, confirming that the transcoders successfully
capture significant MLP functionality.

Additionally, we computed the KL divergence between the logits of the
modified models and the original model:

\begin{table}[h]
  \centering
  \begin{tabular}{lcc}
    \toprule
    & $\mathrm{KL}(\mathrm{Original} \| \mathrm{Transcoder})$ &
    $\mathrm{KL}(\mathrm{Original} \| \mathrm{No\ MLP})$ \\
    \midrule
    Mean & 2.406 & 10.52 \\
    \bottomrule
  \end{tabular}
  \caption{KL divergence between original model and modified models.}
  \label{tab:kl_divergence}
\end{table}

\begin{figure}[h]
  \centering
  \includegraphics[width=0.4\textwidth]{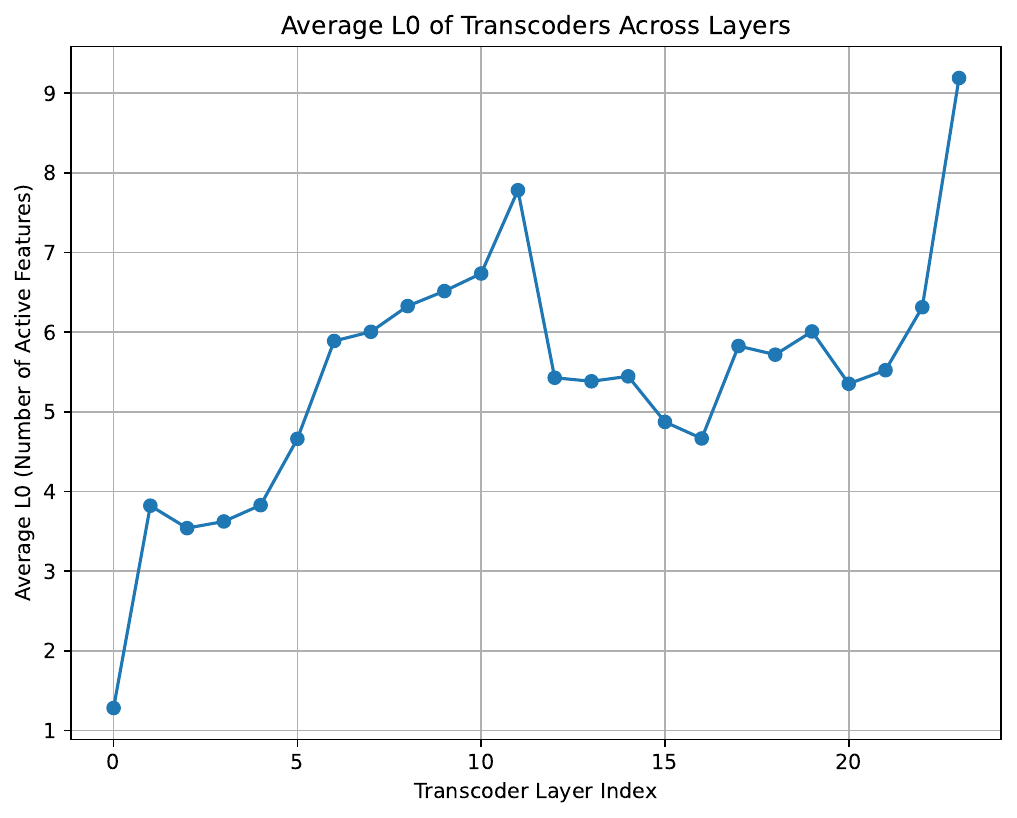}
  \caption{Average L0 values (number of active features per token)
  across transcoder layers.}
  \label{fig:l0_values}
\end{figure}

The average number of activated transcoder features per token (L0
value) across layers is shown in Figure \ref{fig:l0_values}.

\subsection{Human Evaluation of Transcoder Features' Interpretability}

\begin{figure}[h]
  \centering
  \includegraphics[width=0.5\textwidth]{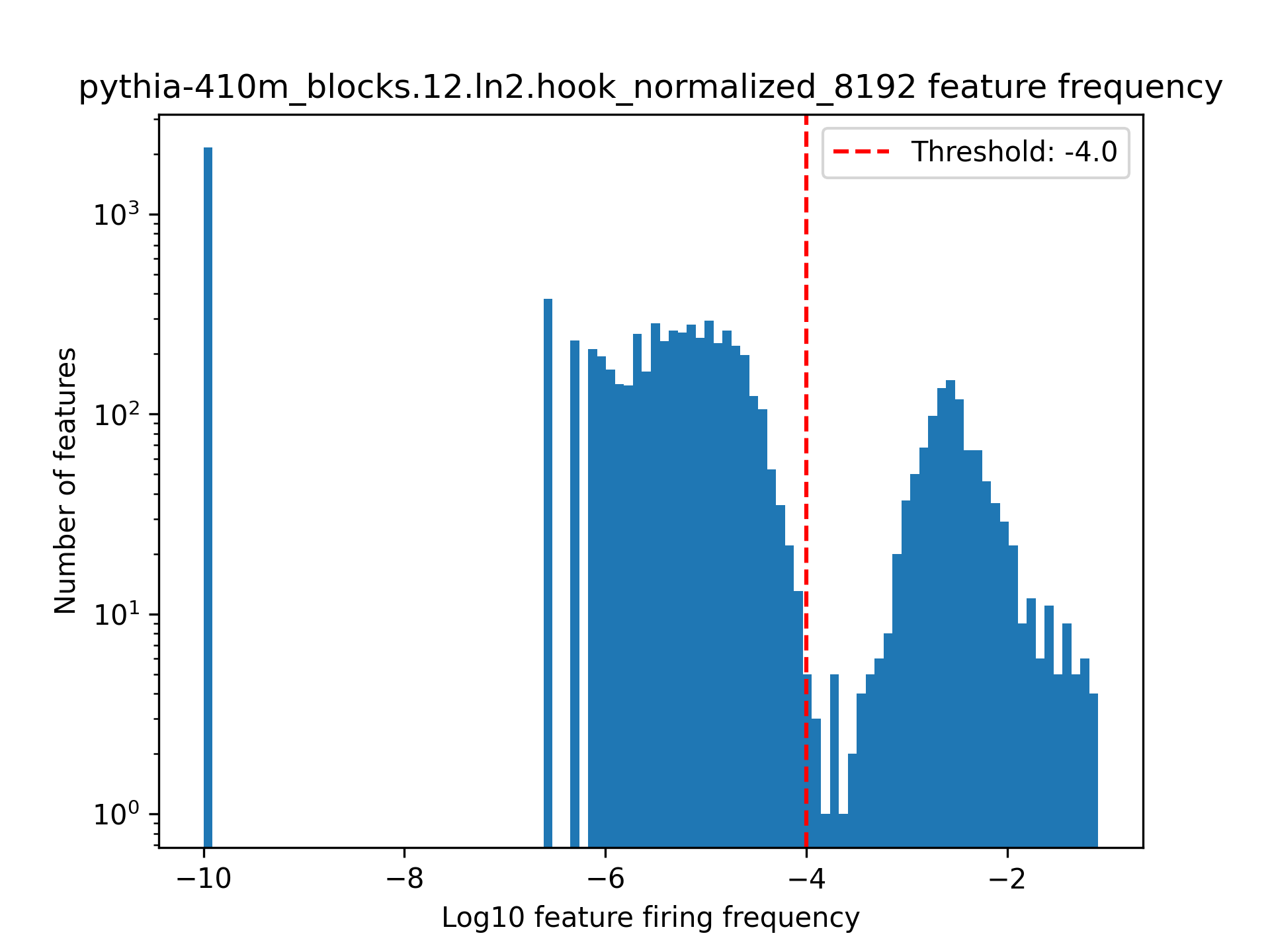}
  \caption{Distribution of live features in layer 12 transcoder with
  $\log_{10} E(f) \geq -4$.}
  \label{fig:live_features}
\end{figure}

To evaluate the interpretability of learned transcoder features, we
conducted a human evaluation study. We focused on transcoder features
from layer 12, defining "live features" as those with $\log_{10} E(f)
\geq -4$, where $E(f)$ represents the probability that feature $f$
activates per token. Figure \ref{fig:live_features} shows the
distribution of live features.

From these live features, we randomly selected 20 features for
detailed interpretation. Features were analyzed by investigating
which tokens activate them. A feature was classified as "gene-level
interpretable" if it consistently activated on tokens corresponding
to specific genes or gene families.

Our evaluation revealed that 7 out of 20 features (35\%) were
gene-level interpretable. Table \ref{tab:feature_interpretation}
presents the detailed analysis of each feature.

\begin{table}[h]
  \centering
  \small
  \begin{tabular}{llp{4cm}c}
    \toprule
    Feature ID & $\log_{10} E(f)$ & Activating Tokens & Gene-Level \\
    & & & Interpretable \\
    \midrule
    6027 & -2.66 & \texttt{\_PSD} token in \texttt{PSD3} gene & Yes \\
    2123 & -2.60 & \texttt{OL} token in \texttt{GOLGA4},
    \texttt{GOLGA8A}, \texttt{GOLPH3} & No \\
    4942 & -1.25 & Trailing \texttt{2} token in genes & No \\
    2459 & -3.01 & \texttt{\_Y} tokens (non-specific) & No \\
    7892 & -2.50 & \texttt{\_ME} token in \texttt{MEIS2}, \texttt{MEF2A} & No \\
    7125 & -2.53 & \texttt{PN} token in \texttt{PTPN} family genes & Yes \\
    3266 & -2.90 & \texttt{SH} token in \texttt{TSHZ2}, \texttt{TSHZ3} & No \\
    1702 & -2.43 & \texttt{AM} token in \texttt{LAM*} genes
    (\texttt{LAMC1}, \texttt{LAMTOR4}) & No \\
    3546 & -2.34 & \texttt{X} token in \texttt{YBX1}, \texttt{YBX3} & Yes \\
    4319 & -2.96 & \texttt{AA} token in \texttt{HSP90AA1} gene & Yes \\
    2709 & -2.57 & \texttt{BL} token in \texttt{ABLIM1},
    \texttt{ABL1}, \texttt{ABLIM3} & No \\
    1283 & -1.71 & \texttt{20} token in \texttt{ZBTB20} gene & No \\
    5085 & -2.60 & \texttt{NA} token in \texttt{GNA*} genes
    (\texttt{GNAI1}, \texttt{GNAI2}, \texttt{GNA14}) & No \\
    2271 & -2.67 & \texttt{NK} token in \texttt{CSNK} family genes & Yes \\
    1980 & -2.91 & \texttt{\_NA} token in \texttt{NAALADL2},
    \texttt{NAIP} & No \\
    2808 & -2.53 & \texttt{OCK} token in \texttt{ROCK1}, \texttt{ROCK2} & No \\
    4619 & -2.90 & \texttt{NN} token in \texttt{TNNI3}, \texttt{TNNC1} & No \\
    5951 & -2.41 & \texttt{O} token in \texttt{FOXO} family genes & Yes \\
    4819 & -2.31 & \texttt{SB} token in \texttt{WSB} family genes & Yes \\
    5280 & -1.98 & \texttt{3} token in \texttt{RPS3}, \texttt{RPS3A} & No \\
    \bottomrule
  \end{tabular}
  \caption{Human interpretation of randomly selected transcoder
    features from layer 12. Note: In token representations, underscore
  (\_) denotes a space character.}
  \label{tab:feature_interpretation}
\end{table}

\subsection{Case Study: Interpreting Cell Type Classification}

To demonstrate the practical application of circuit tracing in scFMs,
we analyzed how the C2S model performs cell type classification. We
extracted a cell labeled as endothelial cell from Heart Cell Atlas v2
and prepared the prompt shown in Figure \ref{fig:prompt}.

\begin{figure}[H]
  \centering
  \begin{minipage}{0.95\textwidth}
    \small
    \texttt{KIAA1217 VWF MT-CO1 FOXN3 MT-CO2 ENG MGLL MT-ND4 MAGI1
      MT-CO3 MT-CYB IQGAP1 SYNE1 CD36 RASAL2 SPARCL1 ST6GALNAC3
      LINC00486 RAPGEF1 ID1 RBMS3 NFIB PTPRB LRMDA ARID2 MT-ATP6
      SMAD2 ZBTB20 RGCC PLAA SLC48A1 TACC1 MECOM RB1 TSPAN14 FRMD4A
      AFDN ANO2 SHOC2 CDC42BPA RASGRF2 CCDC85A ESR2 SLC1A1 FRYL
      MALAT1 FAM241A DIAPH2 TSPAN15 LPAR6 HIF3A ITGA6 PARP14 NSD3
      WNT2B FTX ART4 FBXW11 MTHFR AFF1 KHDRBS1 ZBTB46 ANKRD13C RDX
      SRSF11 ROCK2 SRSF10 BPTF GRB10 ATG4C AHCYL2 CFDP1 AC108449.1
      NUTM2B-AS1 CNTNAP3B FAM214A LNX1 LRP5 MAP2K5 AL138828.1 SIPA1L3
      UVRAG FLI1 GNAI3 EGFL7 ARL17B IL17RA ASAP1 ZFP36L1 PHLPP1 ASXL2
      MT-ND1 NUTM2A-AS1 KIAA1671 LIFR FSD1L PRMT9 DIPK2B KIAA1328
      TIMP3 CEP68 KIAA0100 KALRN PTPRM PPP3CC NR3C1 FBXL7 WDR60
      ABLIM3 WWP1 ZNF761 LINC01060 ZNF274 TTC28 EPAS1 C6ORF89 B2M
      FOXN2 NRP1 SMIM17 VAV3 NCOA3 CCN2 GNAQ WNK1 GMDS ZBTB16 MPPED2.\\
    The corresponding cell type is:}
  \end{minipage}
  \caption{Prompt for cell type classification consisting of 128
  genes ordered by C2S gene rank encoding.}
\end{figure}

This prompt consists of the top 128 genes arranged according to C2S's
gene rank encoding (a cell sentence), followed by a query for cell
type prediction. The C2S model
(\texttt{vandijklab/C2S-Pythia-410m-cell-type-prediction})
successfully predicted "endothelial cell of artery" as the cell type.

\begin{figure}[h]
  \centering
  \begin{subfigure}{0.45\textwidth}
    \centering
    \includegraphics[width=\linewidth]{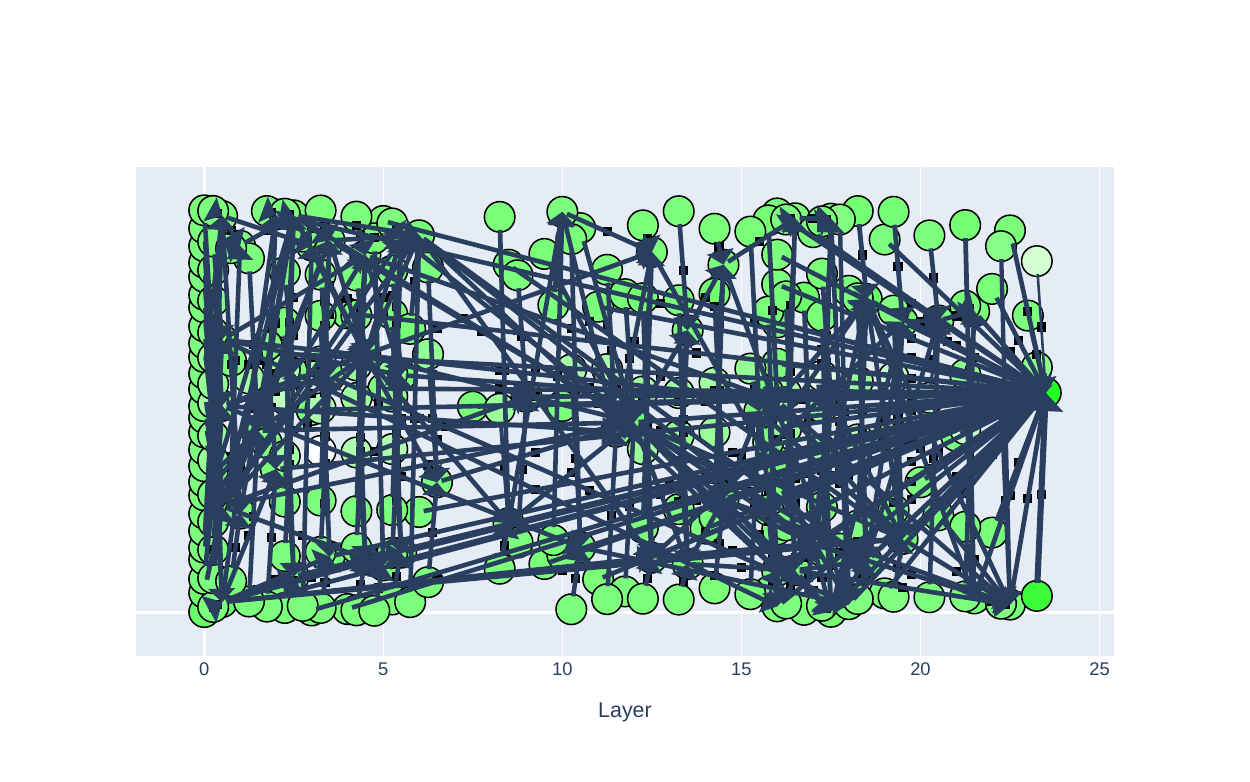}
    \caption{Extracted circuit for feature 3353 activated during endothelial cell classification. The circuit shows the computational graph tracing back from the final prediction token.}
    \label{fig:circuit_endothelial}
  \end{subfigure}\hfill
  \begin{subfigure}{0.50\textwidth}
    \centering
    \includegraphics[width=\linewidth]{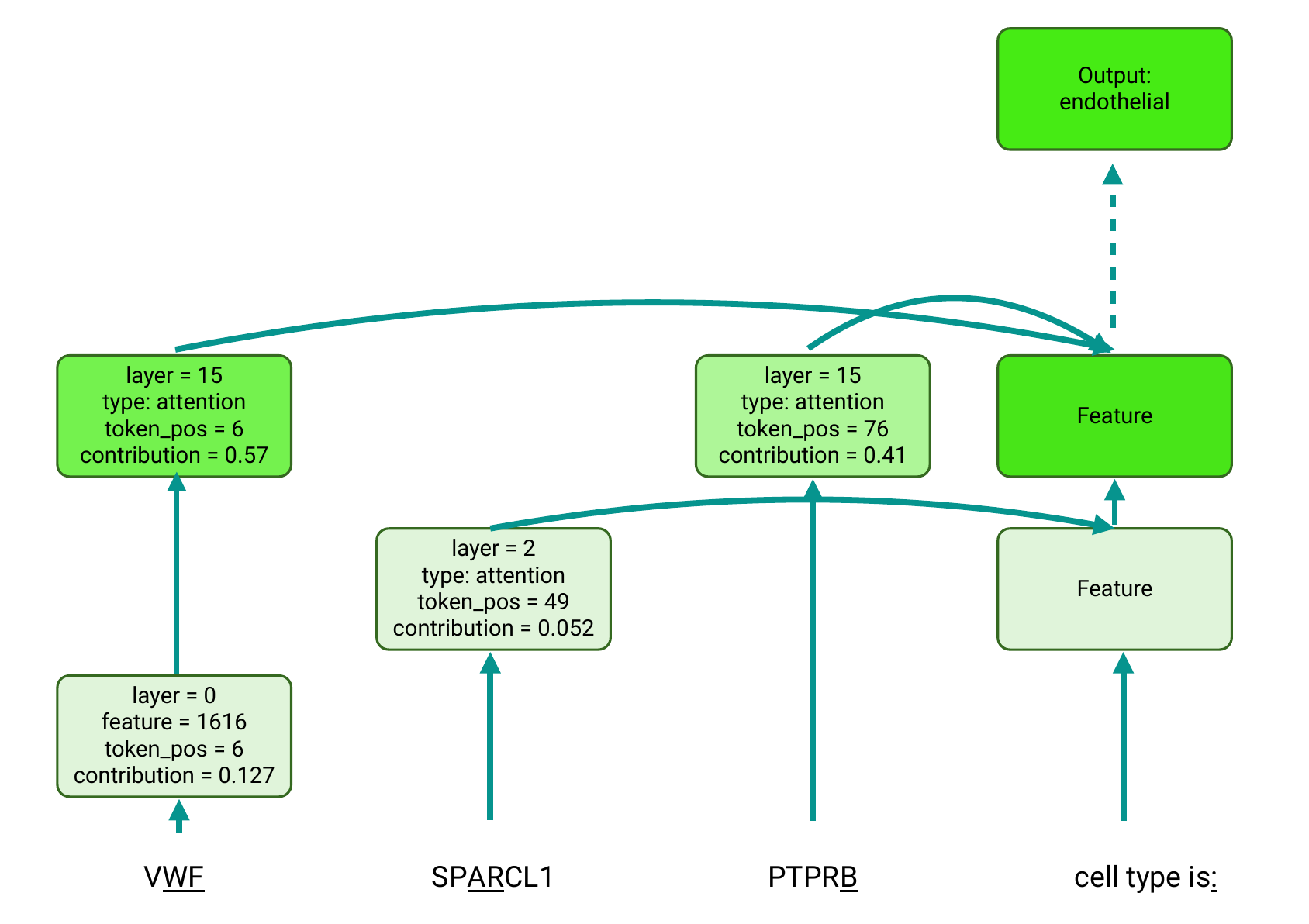}
    \caption{Simplified view of the extracted circuit highlighting gene-related features. The circuit indicates strong contributions from token features corresponding to endothelial-related genes such as VWF, PTPRB, and SPARCL1.}
    \label{fig:circuit_highlighted}
  \end{subfigure}
  \caption{Circuit extraction for endothelial cell classification. (a) Detailed extracted circuit for feature 3353. (b) Simplified circuit emphasizing gene-related features (e.g., VWF, PTPRB, SPARCL1).}
  \label{fig:endothelial_circuits}
\end{figure}

We extracted circuits for features that strongly activated on the
final token (":") in the last layer. Figure
\ref{fig:endothelial_circuits} shows an example circuit extracted for
feature ID 3353, one of the most strongly activated features.

While the extracted circuit contains many activated features, many
correspond to tokens in the text prompt rather than gene names.
Focusing on features activated by gene name tokens, we identified
particularly strong activations for:
\begin{itemize}
  \item VWF (von Willebrand factor): a canonical endothelial marker localized to Weibel--Palade bodies~\cite{Valentijn2011WPB}.
  \item PTPRB (VE-PTP): an endothelial-enriched receptor-type tyrosine phosphatase that regulates junctional integrity and TIE2 signaling~\cite{Drexler2019VEPTP}.
  \item SPARCL1 (hevin): highly expressed in quiescent endothelial cells; contributes to vessel stability and barrier maintenance~\cite{Regensburger2021IBD}.
\end{itemize}

These genes are all closely associated with endothelial cell biology,
suggesting that the extracted circuit captures biologically
meaningful patterns. However, the circuit remains large and complex,
with many features difficult to interpret biologically. This
highlights the need for more refined circuit extraction methods and
feature interpretation techniques in future work.

\section{Related Work}

While mechanistic interpretability of large-scale models including
LLMs has attracted significant attention
\cite{transcoder_circuits_nips2024,kamath2025tracing,huben2024sparse},
the field remains in its early exploratory stages. Particularly,
applications of mechanistic interpretability techniques from natural
language processing to bioinformatics models such as single-cell
analysis models are still limited. Here we summarize the relationship
between our work and these pioneering studies.

Schuster's scFeatureLens
\cite{schuster2025sparseautoencodersmakesense} and work by Claye et
al. \cite{claye2025framework} have applied sparse autoencoders to
scFMs such as Geneformer \cite{Theodoris2023} and scGPT
\cite{Cui2024}, providing frameworks to mechanistically interpret SAE
features as biological concepts. Our research extends this line of
work by utilizing transcoders, an advanced variant of SAEs, to
extract internal decision circuits from scFMs and demonstrate their
correspondence with biological concepts. The frameworks developed in
these prior studies could potentially be applied to individual
transcoder features, suggesting valuable directions for future research.

Additionally, mechanistic interpretability techniques have been
applied to models that directly process genome sequences. For
instance, Brixi et al. developed Evo 2 \cite{evo2}, a genomic
foundation model, and as part of their research incorporated SAE
analysis to investigate the model's internal representations,
revealing that it recognizes biologically important sequence patterns
such as intron-exon boundaries. These studies collectively
demonstrate the growing potential of mechanistic interpretability
methods in understanding biological foundation models.

\section{Conclusion and Future Work}

In this work, we demonstrated that applying transcoders to scFMs
enables the extraction of biologically interpretable features and
internal decision-making circuits. To our knowledge, this is the
first application of transcoders and circuit tracing to mechanistic
interpretability of scFMs, opening new avenues for interpretability
research in single-cell analysis models.

However, this work has several limitations that present opportunities
for future research. First, the biological interpretation of
extracted internal circuits was performed manually, which is
time-consuming and potentially subjective. Future work should develop
automated interpretation methods that can systematically map
transcoder features to biological concepts. Second, the choice of
training dataset significantly impacts the features learned by
transcoders. While we used Heart Cell Atlas v2, exploring other
datasets could reveal more general features and circuits applicable
across different cell types and biological contexts.

Future research directions include developing novel methods to
further enhance scFM interpretability and applying these techniques
to discover new biological insights. The improvement of scFM
interpretability holds significant potential for advancing
single-cell biology, as understanding how these models make decisions
could reveal previously unknown biological relationships and mechanisms.

We anticipate that the field of scFM interpretability will continue
to evolve, bringing new insights to single-cell analysis and
contributing to our understanding of cellular biology at
unprecedented resolution. As scFMs become increasingly integrated
into biological research workflows, ensuring their interpretability
will be crucial for building trust and enabling scientific discovery.

\bibliographystyle{plain}
\bibliography{reference}

\begin{thebibliography}{10}

\bibitem{Biderman2023Pythia}
Stella Biderman, Hailey Schoelkopf, Quentin Anthony, Herbie Bradley, Kyle
  O'Brien, Eric Hallahan, Mohammad~Aflah Khan, Shivanshu Purohit, USVSN~Sai
  Prashanth, Edward Raff, Aviya Skowron, Lintang Sutawika, and Oskar van~der
  Wal.
\newblock Pythia: A suite for analyzing large language models across training
  and scaling.
\newblock In {\em Proceedings of the 40th International Conference on Machine
  Learning}, volume 202 of {\em Proceedings of Machine Learning Research}.
  PMLR, 2023.

\bibitem{evo2}
Garyk Brixi, Matthew~G. Durrant, Jerome Ku, Michael Poli, Greg Brockman, Daniel
  Chang, Gabriel~A. Gonzalez, Samuel~H. King, David~B. Li, Aditi~T. Merchant,
  Mohsen Naghipourfar, Eric Nguyen, Chiara Ricci-Tam, David~W. Romero, Gwanggyu
  Sun, Ali Taghibakshi, Anton Vorontsov, Brandon Yang, Myra Deng, Liv Gorton,
  Nam Nguyen, Nicholas~K. Wang, Etowah Adams, Stephen~A. Baccus, Steven
  Dillmann, Stefano Ermon, Daniel Guo, Rajesh Ilango, Ken Janik, Amy~X. Lu,
  Reshma Mehta, Mohammad~R.K. Mofrad, Madelena~Y. Ng, Jaspreet Pannu,
  Christopher R{\'e}, Jonathan~C. Schmok, John~St. John, Jeremy Sullivan, Kevin
  Zhu, Greg Zynda, Daniel Balsam, Patrick Collison, Anthony~B. Costa, Tina
  Hernandez-Boussard, Eric Ho, Ming-Yu Liu, Thomas McGrath, Kimberly Powell,
  Dave~P. Burke, Hani Goodarzi, Patrick~D. Hsu, and Brian~L. Hie.
\newblock Genome modeling and design across all domains of life with evo 2.
\newblock {\em bioRxiv}, 2025.

\bibitem{claye2025framework}
Charlotte Claye, Pierre Marschall, Wassila Ouerdane, C{\'e}line Hudelot, and
  Julien Duquesne.
\newblock A framework to extract and interpret biological concepts from
  scrnaseq generative foundation models.
\newblock In {\em ICML 2025 Generative AI and Biology (GenBio) Workshop}.

\bibitem{Cui2024}
Haotian Cui, Chloe Wang, Hassaan Maan, Kuan Pang, Fengning Luo, Nan Duan, and
  Bo~Wang.
\newblock scgpt: toward building a foundation model for single-cell multi-omics
  using generative ai.
\newblock {\em Nature Methods}, 21(8):1470–1480, February 2024.

\bibitem{Drexler2019VEPTP}
Hannes C.~A. Drexler and {others}.
\newblock Vascular endothelial receptor tyrosine phosphatase: Identification of
  novel substrates related to junctions and a ternary complex with ephb4 and
  tie2.
\newblock {\em Molecular \& Cellular Proteomics}, 18(10):2058--2077, 2019.

\bibitem{transcoder_circuits_nips2024}
Jacob Dunefsky, Philippe Chlenski, and Neel Nanda.
\newblock Transcoders find interpretable llm feature circuits.
\newblock In {\em Proceedings of the 38th International Conference on Neural
  Information Processing Systems}, NIPS '24, Red Hook, NY, USA, 2025. Curran
  Associates Inc.

\bibitem{Elhage2022ToyModels}
Nelson Elhage, Tristan Hume, Catherine Olsson, Nicholas Schiefer, Tom Henighan,
  Shauna Kravec, Zac Hatfield-Dodds, Robert Lasenby, Dawn Drain, Carol Chen,
  Roger Grosse, Sam McCandlish, Jared Kaplan, Dario Amodei, Martin Wattenberg,
  and Christopher Olah.
\newblock Toy models of superposition.
\newblock {\em arXiv preprint arXiv:2209.10652}, 2022.

\bibitem{Elhage2021Framework}
Nelson Elhage, Neel Nanda, Catherine Olsson, Tom Henighan, Nicholas Joseph, Ben
  Mann, Amanda Askell, Yuntao Bai, Anna Chen, Tom Conerly, Nova DasSarma, Dawn
  Drain, Deep Ganguli, Zac Hatfield-Dodds, Danny Hernandez, Andy Jones, Jackson
  Kernion, Liane Lovitt, Kamal Ndousse, Dario Amodei, Tom Brown, Jack Clark,
  Jared Kaplan, Sam McCandlish, and Chris Olah.
\newblock A mathematical framework for transformer circuits.
\newblock Transformer Circuits Thread, 2021.
\newblock Accessed 2025-09-08.

\bibitem{huben2024sparse}
Robert Huben, Hoagy Cunningham, Logan~Riggs Smith, Aidan Ewart, and Lee
  Sharkey.
\newblock Sparse autoencoders find highly interpretable features in language
  models.
\newblock In {\em The Twelfth International Conference on Learning
  Representations}, 2024.

\bibitem{kamath2025tracing}
Harish Kamath, Emmanuel Ameisen, Isaac Kauvar, Rodrigo Luger, Wes Gurnee, Adam
  Pearce, Sam Zimmerman, Joshua Batson, Thomas Conerly, Chris Olah, and Jack
  Lindsey.
\newblock Tracing attention computation: Attention connects features, and
  features direct attention.
\newblock {\em Transformer Circuits Thread}, 2025.

\bibitem{heart_cell_atlas_v2}
Kazumasa Kanemaru, James Cranley, Daniele Muraro, Antonio M~A Miranda, Siew~Yen
  Ho, Anna Wilbrey-Clark, Jan Patrick~Pett, Krzysztof Polanski, Laura
  Richardson, Monika Litvinukova, Natsuhiko Kumasaka, Yue Qin, Zuzanna
  Jablonska, Claudia~I Semprich, Lukas Mach, Monika Dabrowska, Nathan Richoz,
  Liam Bolt, Lira Mamanova, Rakeshlal Kapuge, Sam~N Barnett, Shani Perera,
  Carlos Talavera-L{\'o}pez, Ilaria Mulas, Krishnaa~T Mahbubani, Liz Tuck,
  Lu~Wang, Margaret~M Huang, Martin Prete, Sophie Pritchard, John Dark, Kourosh
  Saeb-Parsy, Minal Patel, Menna~R Clatworthy, Norbert H{\"u}bner, Rasheda~A
  Chowdhury, Michela Noseda, and Sarah~A Teichmann.
\newblock Spatially resolved multiomics of human cardiac niches.
\newblock {\em Nature}, 619(7971):801--810, July 2023.

\bibitem{Levine2024C2S}
Daniel Levine, Syed~A Rizvi, Sacha L{\'e}vy, Nazreen Pallikkavaliyaveetil,
  David Zhang, Xingyu Chen, Sina Ghadermarzi, Ruiming Wu, Zihe Zheng, Ivan
  Vrkic, Anna Zhong, Daphne Raskin, Insu Han, Antonio~Henrique
  De~Oliveira~Fonseca, Josue Ortega~Caro, Amin Karbasi, Rahul~Madhav Dhodapkar,
  and David Van~Dijk.
\newblock Cell2sentence: Teaching large language models the language of
  biology.
\newblock In {\em Proceedings of the 41st International Conference on Machine
  Learning}, volume 235 of {\em Proceedings of Machine Learning Research},
  pages 27299--27325. PMLR, 2024.

\bibitem{Regensburger2021IBD}
Daniela Regensburger and {others}.
\newblock Matricellular protein sparcl1 regulates blood vessel integrity and
  antagonizes inflammatory bowel disease.
\newblock {\em Inflammatory Bowel Diseases}, 27(9):1491--1502, 2021.

\bibitem{schuster2025sparseautoencodersmakesense}
Viktoria Schuster.
\newblock Can sparse autoencoders make sense of gene expression latent variable
  models?, 2025.

\bibitem{Theodoris2023}
Christina~V. Theodoris, Ling Xiao, Anant Chopra, Mark~D. Chaffin, Zeina~R.
  Al~Sayed, Matthew~C. Hill, Helene Mantineo, Elizabeth~M. Brydon, Zexian Zeng,
  X.~Shirley Liu, and Patrick~T. Ellinor.
\newblock Transfer learning enables predictions in network biology.
\newblock {\em Nature}, 618(7965):616–624, May 2023.

\bibitem{Valentijn2011WPB}
Karine~M. Valentijn, J.~Evan Sadler, Jack~A. Valentijn, Jan Voorberg, and
  Jeroen Eikenboom.
\newblock Functional architecture of weibel-palade bodies.
\newblock {\em Blood}, 117(19):5033--5043, 2011.

\bibitem{HF_C2S_Pythia_410m_CellTypePred}
{van Dijk Lab}.
\newblock vandijklab/c2s-pythia-410m-cell-type-prediction.
\newblock Hugging Face, 2025.
\newblock Model card. Trained on $\sim$57M single cells from CellxGene and HCA.
  Accessed 2025-09-08.

\end{thebibliography}

\end{document}